\documentclass[11pt]{article}

\usepackage[a4paper,margin=1in]{geometry}
\usepackage[utf8]{inputenc}
\usepackage[T1]{fontenc}
\usepackage{lmodern}
\usepackage{microtype}
\usepackage{amsmath,amssymb,amsthm}
\usepackage{booktabs}
\usepackage{array}
\usepackage{multirow}
\usepackage{graphicx}
\usepackage[table,xcdraw]{xcolor}
\usepackage{colortbl}
\usepackage{algorithm}
\usepackage{algpseudocode}
\usepackage{caption}
\usepackage{subcaption}
\usepackage{enumitem}
\usepackage{url}
\usepackage{tikz}
\usetikzlibrary{positioning,shapes.geometric,arrows.meta,fit,backgrounds,calc}
\usepackage{placeins}
\usepackage[round,authoryear]{natbib}
\usepackage{hyperref}

\hypersetup{
    colorlinks=true,
    linkcolor=blue!50!black,
    citecolor=blue!50!black,
    urlcolor=blue!50!black
}

\newtheorem{theorem}{Theorem}

\definecolor{rowRed}{rgb}{1.00,0.92,0.92}
\definecolor{rowYellow}{rgb}{1.00,0.97,0.84}
\definecolor{rowGreen}{rgb}{0.90,0.97,0.90}
\definecolor{rowOrange}{rgb}{1.00,0.92,0.84}
\definecolor{rowBlue}{rgb}{0.92,0.95,1.00}
\definecolor{rowPurple}{rgb}{0.94,0.92,1.00}
\definecolor{rowGray}{rgb}{0.94,0.94,0.94}

\title{Iterative Causal Discovery: Per-Edge Impossibility Certificates,
Tier-Aware Oracle Queries, and the $1{+}K$ Lower Bound}

\author{%
  Eichi Uehara\\
  aflo, Inc., Tokyo, Japan\\
  \texttt{eichi.uehara@aflo.one}
}

\date{\today}

\begin{document}
\maketitle

\begin{abstract}
Causal-discovery algorithms return a directed graph, yet provide
no principled means of distinguishing edge directions identified
by the data from those assigned without an identifying assumption.
Under the standard Markov and faithfulness conditions, the
observational distribution identifies only a Markov equivalence
class; orientations within that class are not determined by the
joint distribution and cannot be recovered from additional samples
alone, but require either a functional restriction or an
intervention. We introduce a protocol for observational causal
discovery on continuous data that attaches to each candidate edge
a discrete \emph{impossibility certificate}: a
\textsc{resolved\_*} code records the identifiability theorem
under which the direction was committed, while an
\textsc{impossible\_*} code records the failure mode together
with the specific question a domain expert must answer to resolve
it. The bivariate cascade is extended with five gated
identifiability tiers---LSNM, IGCI, Stein, MDL, and PEIT---that
abstain when their precondition test rejects. Two oracle
primitives, the meta-hub query and the node-children query,
jointly establish an upper bound of $1{+}K$ expert interactions
sufficient to recover any DAG, where $K$ denotes the number of
non-leaf vertices. Under an ideal-oracle assumption, the bound is
met exactly on the asia, sachs, child, and alarm benchmarks.
\end{abstract}

\section{Introduction}
\label{sec:intro}

Causal structure underpins counterfactual reasoning, intervention
design, and out-of-distribution
generalisation~\citep{pearl2009causality,scholkopf2021toward,peters2017elements},
and its recovery from observational data is a long-standing
objective of machine learning. Existing algorithms scale to graphs
of hundreds of vertices and return a directed graph estimate, but
do not indicate, for each edge, whether the assigned orientation
was identified by the data or imposed by the algorithm in order
to satisfy its output format. The distinction is material:
several widely used continuous-optimisation methods have been
shown to exploit data-generation artefacts such as
varsortability~\citep{reisach2021beware,kaiser2022tearing}, and
without a per-edge record of identification such artefacts are
indistinguishable from genuine identification.

The underlying identifiability obstacle is classical. Under the
standard Markov and faithfulness conditions, the observational
distribution identifies only a Markov equivalence class: orientations of edges
that do not lie on a v-structure are not determined by the joint
distribution, and no algorithm operating on that distribution
alone can recover
them~\citep{glymour2019review,pearl2009causality,spirtes2000causation}.
Breaking the equivalence class requires either an additional
functional restriction---an additive-noise
model~\citep{hoyer2009nonlinear},
LiNGAM~\citep{shimizu2006linear,shimizu2011directlingam},
location-scale noise~\citep{strobl2022identifying,immer2023identifiability},
the information-geometric assumption of
IGCI~\citep{janzing2012information}, Stein-score
identifiability~\citep{rolland2022score}, or the algorithmic
Markov condition~\citep{janzing2010causal}---or an
intervention~\citep{eberhardt2007number,hauser2014two,tigas2022interventions};
increasing the sample size does not.
\citet{mooij2016distinguishing} survey the bivariate
identifiability landscape and document the substantial fraction
of empirical edges that fall outside any single one of these
regimes.

The present analysis is restricted to observational data with
\emph{continuous} variables, for which the residualisation that
underlies the cascade tiers and the multi-tier mediator search is
well-defined. Extension of the certification pipeline to discrete
Bayesian networks constitutes a separate line of future work; the
discrete-BN runs reported in
Section~\ref{sec:experiments:multibench} exercise only the oracle
layer and serve as a robustness check, not as the intended
deployment regime.

Within this scope, a faithful causal-discovery method may, for any
candidate edge $(X, Y)$, return one of three responses:
(i)~the data identifies a direction under a named theorem;
(ii)~the apparent dependence is mediated by an observed variable,
so that the edge is not direct; or (iii)~the data are
insufficient, and a specific prior question would resolve the
orientation. The literature provides
theoretical foundations for
(i)~\citep{hoyer2009nonlinear,shimizu2006linear,zhang2009identifiability,janzing2012information,strobl2022identifying,rolland2022score,hyvarinen2013pairwise},
classical machinery
for~(ii)~\citep{spirtes2000causation,colombo2012learning}, and
three established regimes for~(iii): linear-Gaussian
pairs~\citep{pearl2009causality}, latent confounding, and
finite-sample regressor
misspecification~\citep{mooij2016distinguishing}.
Existing methods pursue distinct design goals
(Figure~\ref{fig:problem}) and consequently fail to separate
these three responses at the edge level. Continuous-optimisation
frameworks (NOTEARS~\citep{zheng2018notears},
DAGMA~\citep{bello2022dagma},
DirectLiNGAM~\citep{shimizu2011directlingam}, and the score-based
PC and GES
methods~\citep{kalisch2007estimating,chickering2002optimal})
scale to large graphs by returning a fully oriented DAG; in doing
so, type-(iii) edges receive a direction selected by the
optimisation objective rather than by the data. Constraint-based
methods (FCI~\citep{spirtes2000causation,zhang2008causal} and its
PAG output) correctly leave type-(iii) edges undirected, but do
not specify the prior input that would resolve each. Bayesian
DAG-posterior methods (DiBS~\citep{lorch2021dibs},
BCD-Nets~\citep{cundy2021bcd},
AVICI~\citep{lorch2022amortized},
VCN~\citep{annadani2021variational}) return per-edge marginal
probabilities, which can be viewed as a soft analogue of an
identifiability statement; the probabilities marginalise over
assumptions rather than identifying which assumption resolves a
given edge. Active-learning
methods~\citep{he2008active,shanmugam2015learning,hauser2014two,tigas2022interventions}
elicit interventions or topological orders, typically at a budget
exceeding what type-(iii) edges alone require. The protocol
proposed below addresses this gap by isolating the type-(iii)
edges and identifying, for each, the question whose answer would
resolve it.

\begin{figure*}[t]
\centering
\begin{tikzpicture}[
  font=\sffamily\footnotesize,
  every node/.style={align=center},
  full/.style={draw,rounded corners=2pt,inner sep=3pt,fill=white,
                minimum width=11.5cm,minimum height=8mm},
  col/.style={draw,rounded corners=2pt,inner sep=3pt,fill=white,
                minimum width=5.5cm,minimum height=10mm,text width=5cm},
  arrow/.style={-Stealth,semithick},
  node distance=4pt,
]
\node[full,fill=rowBlue] (data)  {observational data};
\node[full,below=of data] (skel) {\textbf{Stage 1:} FDR-controlled HSIC skeleton $+$ multi-tier mediator search};
\node[full,below=of skel] (cas)
  {\textbf{Stage 2:} identifiability cascade
    $L_0 \,{\to}\, L_1 \,{\to}\, L_\mathrm{LSNM} \,{\to}\, L_\mathrm{IGCI} \,{\to}\, L_\mathrm{Stein} \,{\to}\, L_\mathrm{MDL} \,{\to}\, L_2 \,{\to}\, L_\mathrm{PEIT}$ \\
    each tier gated by a precondition test};

\node[col,fill=rowGreen,below left=8pt and 4pt of cas.south]   (res)
  {\textbf{\textsc{resolved\_*}}\\\scriptsize data identifies direction\\\scriptsize (commit immediately, $0$ queries)};
\node[col,fill=rowYellow,below right=8pt and 4pt of cas.south] (imp)
  {\textbf{\textsc{impossible\_*}}\\\scriptsize data insufficient: certificate\\\scriptsize names the question to ask};

\node[col,fill=rowOrange,below=of res] (prop)
  {\textbf{auto-propagation}\\\scriptsize acyclicity $+$ Meek-R1 + R3\\\scriptsize $+$ transitive d-separation};
\node[col,fill=rowOrange,below=of imp] (oracle)
  {\textbf{Stage 3: oracle queries}\\\scriptsize per-edge \textsc{fwd}/\textsc{bwd}/\textsc{absent} \emph{or}\\\scriptsize meta-hub $+$ node-children ($1{+}K$)};

\node[full,fill=rowGreen,below=10pt of $(prop.south)!0.5!(oracle.south)$]
  (out) {\textbf{certified DAG} $+$ per-edge certificate $+$ audit trace};

\draw[arrow] (data) -- (skel);
\draw[arrow] (skel) -- (cas);
\draw[arrow] (cas.south) -| (res.north);
\draw[arrow] (cas.south) -| (imp.north);
\draw[arrow] (res) -- (prop);
\draw[arrow] (imp) -- (oracle);
\draw[arrow] (prop.south) |- (out.north -| prop);
\draw[arrow] (oracle.south) |- (out.north -| oracle);
\end{tikzpicture}
\caption{Protocol overview. Stage 2's cascade contains five new
gated safe tiers (LSNM, IGCI, Stein, MDL, PEIT) in addition to the
classical $L_0$/$L_1$/$L_2$ ANM lattice. Each tier abstains when
its precondition test fails, so a candidate that would otherwise
be committed under an incorrect assumption is relabelled
\textsc{impossible\_*}. Stage 3 issues either per-edge queries or
meta-hub $+$ node-children queries.}
\label{fig:method}
\end{figure*}
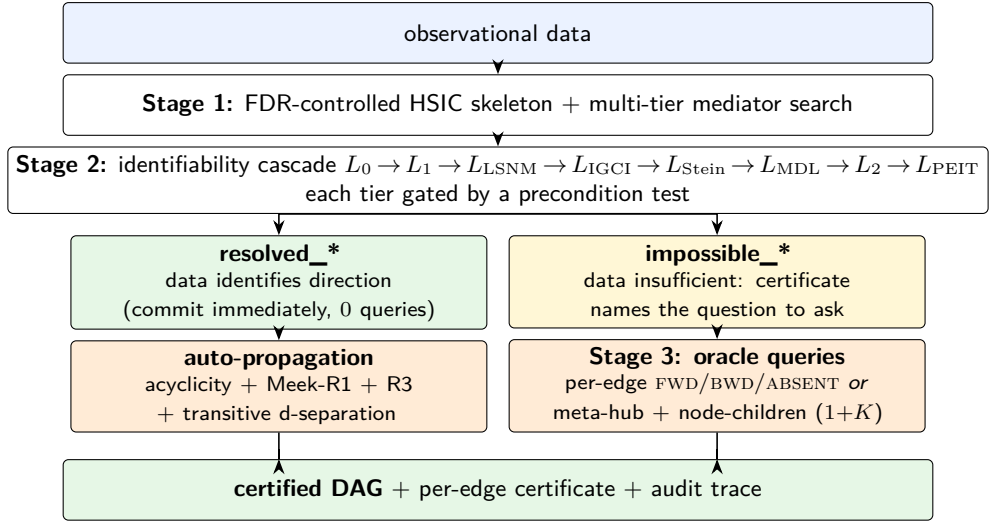

The protocol emits, for each candidate edge of a data-only
skeleton, one label drawn from a fixed discrete code set
(Figure~\ref{fig:method}). Two \textsc{resolved\_*} codes
correspond to response~(i)---the data identifies a direction
under a named theorem---and~(ii)---the edge is mediated. A family
of \textsc{impossible\_*} codes corresponds to response~(iii);
each code records the specific failure mode (linear-Gaussian,
latent-likely, regressor-inconsistent, four regime-specific codes
for circular, binary-continuous, count, and high-cardinality
data, and one cross-tier-disagreement code) together with the
question a domain expert would be required to answer.
The bivariate cascade is extended with five new identifiability
tiers
(LSNM~\citep{strobl2022identifying,immer2023identifiability},
IGCI~\citep{janzing2012information},
Stein~\citep{rolland2022score}, MDL~\citep{janzing2010causal},
and PEIT); each tier is preconditioned by a statistical test, and
a tier whose precondition is rejected \emph{abstains}, demoting
the candidate to an \textsc{impossible\_*} label rather than
issuing an orientation the tier cannot certify. Two oracle
primitives, the \emph{meta-hub query}---a single interaction
returning the top-$K$ network hubs---and the
\emph{node-children query}---a single interaction returning a
node's direct children---permit a hub-aware practitioner to
commit every outgoing edge of a node in a single query.

Relative to the methods reviewed in
Section~\ref{sec:related}, the proposed protocol provides four
guarantees:
\begin{enumerate}[leftmargin=*,itemsep=0pt,topsep=2pt]
\item \textbf{Per-edge auditability.} Every committed orientation
records the theorem under which it was issued
($L_0$, $L_1$, $L_\text{LSNM}$, $L_\text{IGCI}$, $L_\text{Stein}$,
$L_\text{MDL}$, $L_2$, $L_\text{PEIT}$, mediator search, or
oracle); every uncommitted edge records the corresponding failure
mode.
\item \textbf{Abstention in place of forced orientation on
unidentifiable edges.} Each cascade tier abstains when its
precondition is rejected, and the cross-tier disagreement guard
demotes any commit that a higher-precision tier contradicts; a
candidate that would otherwise be force-oriented is relabelled
\textsc{impossible\_*}.
\item \textbf{Targeted prior elicitation.} Only IMPOSSIBLE edges
are presented to the practitioner, each accompanied by the
question that resolves its failure mode. Two interaction types
are supported: a per-edge query returning one of
\textsc{fwd}~/~\textsc{bwd}~/~\textsc{absent}, and a
node-children query that commits all outgoing edges of a node in
a single interaction.
\item \textbf{Information-theoretic floor of $1{+}K$ interactions
under an ideal-oracle assumption.} Provided a domain expert
answers each meta-hub and node-children query correctly, the
protocol recovers the ground-truth DAG at precision $=$
recall~$=1.000$ in exactly $1{+}K$ interactions, where $K$
denotes the number of non-leaf vertices (Theorem~\ref{thm:1k}).
The bound is information-theoretic and holds independently of
the edge count; performance under noisy or probabilistic
elicitation is deferred to future work.
\end{enumerate}

\begin{figure}[t]
\centering
\includegraphics[width=0.85\linewidth]{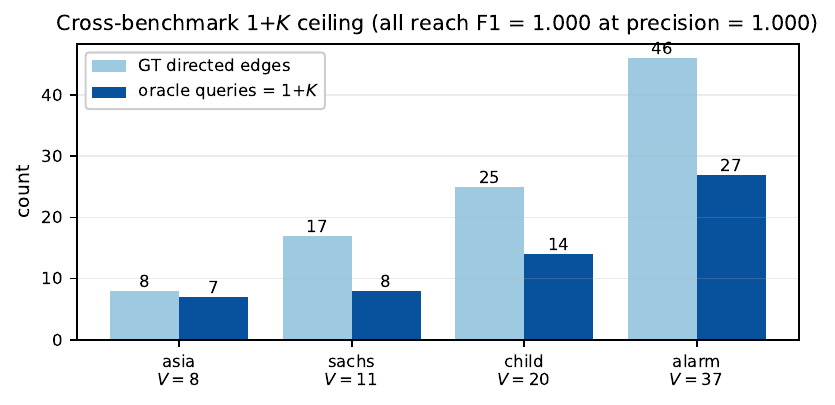}
\caption{Cross-benchmark $1{+}K$ upper bound under the
ideal-oracle assumption. Assuming a domain expert who answers each
meta-hub and node-children query correctly, the protocol recovers
each ground-truth DAG at precision = recall = F1 = $1.000$ using
exactly $1{+}K$ oracle interactions, where $K$ is the count of
non-leaf vertices. Query count grows much slower than edge count
(alarm: $46$ edges committed by $27$ interactions). Performance
under noisy or probabilistic elicitation is future work.}
\label{fig:multibench-1k}
\end{figure}

These guarantees are verified on standard benchmarks (asia,
sachs, child, and alarm; Figure~\ref{fig:multibench-1k}) and on
synthetic data designed to isolate each new tier
(Section~\ref{sec:experiments:safetier-verify}). The protocol is
deterministic given a seed; the largest benchmark
(alarm, $V{=}37$) was reproduced end-to-end on a remote Ray
cluster.

\begin{figure*}[t]
\centering
\small
\begin{tabular}{@{}p{0.20\linewidth}@{\hspace{0.5em}}p{0.74\linewidth}@{}}
\toprule
\rowcolor{rowRed}
\textbf{Forced-DAG}\newline\scriptsize PC, GES, NOTEARS, DAGMA, DirectLiNGAM
& Commit a direction on every edge for scalability $\Rightarrow$
  unidentifiable edges receive a forced direction without a
  per-edge diagnostic of whether data or the optimiser chose it.\\[3pt]
\rowcolor{rowYellow}
\textbf{PAG / FCI}
& Mark ambiguous edges undirected $\Rightarrow$ correct, but no
  targeted question per edge for the practitioner.\\[3pt]
\rowcolor{rowRed}
\textbf{Full-tier /\newline active-learning}
& Ask for a full topological order or interventions
  ($V \log_2 V$ bits up front) $\Rightarrow$ over-asks on the easy
  edges, still under-resolves the hard ones.\\[3pt]
\rowcolor{rowGreen}
\textbf{This paper:\newline per-edge\newline certificates}
& $\textsc{resolved\_*}$: data identifies direction under a named
  theorem, or edge is mediated. $\textsc{impossible\_*}$: data
  insufficient $\Rightarrow$ certificate names the specific question
  to ask.\\
\bottomrule
\end{tabular}
\caption{Existing causal-discovery output formats and the
per-edge certificate alternative. Forced-DAG and PAG methods
target different design goals (scalability and sound
non-identifiability marking, respectively); the proposed protocol
addresses an orthogonal goal by identifying, for each
unidentifiable edge, the question whose answer would resolve
it.}
\label{fig:problem}
\end{figure*}

\section{Related work}
\label{sec:related}

The related literature is organised below into five strands, each
addressing one component of the problem considered in this paper.

Continuous-optimisation methods cast structure learning as a
differentiable problem and return a fully oriented DAG.
NOTEARS~\citep{zheng2018notears}, DAGMA~\citep{bello2022dagma},
and DirectLiNGAM~\citep{shimizu2011directlingam} are
representative; the forced-commitment design favours scalability
but does not expose, per edge, whether the chosen direction was
identified by the data or assigned by the optimiser.

Constraint-based methods are the historical alternative and rely
on conditional-independence testing.
PC~\citep{kalisch2007estimating},
GES~\citep{chickering2002optimal}, and
FCI~\citep{spirtes2000causation,zhang2008causal} return a CPDAG
or PAG, leaving unidentifiable edges undirected. They are sound
on the unidentifiable set, yet do not name the specific prior
input that would resolve each undirected edge.

A third line treats the DAG as a random variable and learns a
posterior over structures. DiBS~\citep{lorch2021dibs},
BCD-Nets~\citep{cundy2021bcd}, AVICI~\citep{lorch2022amortized},
and VCN~\citep{annadani2021variational} return per-edge marginal
posterior probabilities, which can be read as a soft analogue of
an impossibility certificate. Our protocol differs by emitting a
discrete, theorem-tagged code per edge that names the failure
mode and the practitioner question, rather than a probability
that marginalises over assumptions.

A parallel literature bounds the number of \emph{interventions}
required to identify a DAG.
\citet{eberhardt2007number}, \citet{hauser2014two}, and
\citet{tigas2022interventions,shanmugam2015learning,he2008active}
provide such budgets; the node-children query proposed in this
paper is a non-interventional analogue, in which the practitioner
enumerates the direct children of a node rather than performing an
intervention upon it. The $1{+}K$ count established in
Section~\ref{sec:experiments} is the elicitation-side counterpart
of those intervention budgets.

The use of expert input to refine causal structure has a long
history. \citet{heckerman1995learning} pioneered prior elicitation
for Bayesian-network learning, and
\citet{russo2023argumentation} introduce an argumentation
framework for interactive causal discovery. The present work
extends this line by combining the bivariate identifiability
cascade with the per-edge certificate, so that elicited prior
input is requested only at the edges that require it, and only
through the question that resolves the certificate code.

\section{Methodology}
\label{sec:method}

Given an observational dataset $X \in \mathbb{R}^{N \times V}$,
the protocol emits, for every candidate edge in a data-only
skeleton, one label drawn from a fixed code set:
\textsc{resolved\_*} when the data identifies the direction (or
the edge is mediated), and \textsc{impossible\_*} when the data
is insufficient, in which case the label names the question the
practitioner must answer. The protocol comprises three stages:
(1)~skeleton construction and multi-tier mediator search;
(2)~an eight-tier identifiability cascade with precondition-gated
commits; and (3)~targeted oracle queries on the IMPOSSIBLE
residual, interleaved with auto-propagation after each commit.
The protocol is deterministic given a seed. The cascade-then-oracle
composition is presented below as a query-minimisation algorithm,
followed by a description of the data-only stage and of the
oracle interaction layer.

\label{sec:method:algorithm}%
Every practitioner interaction consumes prior information, so the
algorithm minimises interactions through three layers.
Layer~1 commits as many edges as possible from the data alone,
without issuing a query. Layer~2 attaches a regime-specific code
to each remaining IMPOSSIBLE edge, reducing the information cost
of the corresponding question. Layer~3 selects, for each residual
edge, the oracle interaction of lowest cost.
Table~\ref{tab:mechanisms} enumerates the mechanisms by layer.

\begin{table}[h]
\centering\small
\caption{The fifteen query-minimisation mechanisms, grouped by
layer. Layer~1 issues no queries; Layer~2 reduces the
information cost of the remaining queries by labelling each
IMPOSSIBLE edge with a regime-specific code; Layer~3 selects, for
each residual edge, the oracle interaction of lowest cost.}
\label{tab:mechanisms}
\begin{tabular}{l p{8.5cm}}
\toprule
\multicolumn{2}{l}{\textbf{Layer 1: observation-time auto-resolution (no queries issued).}}\\
\midrule
M1 & FDR-controlled HSIC skeleton; drops marginal-independent pairs.\\
M2 & Multi-tier mediator search (tier-1 single-$Z$, tier-2 pairs, tier-3 Markov blanket).\\
M3 & Eight-level cascade $L_0 \to L_1 \to L_\text{LSNM} \to L_\text{IGCI} \to L_\text{Stein} \to L_\text{MDL} \to L_2 \to L_\text{PEIT}$. Each tier commits direction from data when its precondition test passes (\S\ref{sec:method:cascade}).\\
M4 & Per-tier precondition gates (Shapiro--Wilk, integer-valued, heteroscedasticity, noise-variance ratio). A failing gate abstains $\Rightarrow$ wrong commit converted to \textsc{impossible\_*}.\\
M5 & $L_0$-disagreement guard: demotes $L_0$'s commit to \textsc{impossible\_l0\_disagrees\_with\_high\_tier} when $L_\text{IGCI}$, $L_\text{Stein}$, or $L_\text{MDL}$ pick the opposite.\\
M6 & Bivariate-confirmed auto-propagation (acyclicity, Meek R1, Meek R3) gated by a $\ge 2{\times}$ bivariate confirm-ratio.\\
M7 & Parent-conditioned ANM re-audit after each oracle commit; optional safe-tier variant runs the cascade on the residual stream.\\
M8 & Transitive d-separation: IMPOSSIBLE edges re-tested for independence given the growing committed-parents set.\\
\midrule
\multicolumn{2}{l}{\textbf{Layer 2: regime-specific IMPOSSIBLE labels (lower-cost queries).}}\\
\midrule
M9 & Six classical codes: \textsc{impossible\_r1}, \textsc{impossible\_latent\_likely}, \textsc{impossible\_regressor\_inconsistent}, \textsc{impossible\_nonlinear\_weak}, \textsc{impossible\_hoc\_ambiguous}, \textsc{impossible\_ambiguous}.\\
M10 & Five new codes added here: \textsc{impossible\_l0\_disagrees\_with\_high\_tier} (M5) and four regime detectors $\{$\textsc{circular}, \textsc{binary\_continuous}, \textsc{count}, \textsc{high\_cardinality\_discrete}$\}$, each carrying a regime-specific practitioner question.\\
\midrule
\multicolumn{2}{l}{\textbf{Layer 3: oracle interaction primitives (selecting the most informative query).}}\\
\midrule
M11 & Per-edge query: practitioner returns one of \textsc{fwd} /
\textsc{bwd} / \textsc{absent} for a single candidate edge
($\log_2 3$ bits of information).\\
M12 & Information-value query ordering: $\mathrm{value}(e) = \min(\mathrm{prop\_count}(\textsc{fwd}), \mathrm{prop\_count}(\textsc{bwd}))$ (default; expected-value variant available).\\
M13 & \emph{Meta-hub query}: one interaction returns the top-$K$ out-degree nodes of the network.\\
M14 & \emph{Node-children query}: one interaction returns the direct children of a single node, committing all outgoing edges. Combined with M13, achieves the $1{+}K$ ceiling (Theorem~\ref{thm:1k}).\\
M15 & Missing-edge recovery with three optional filters (marginal-HSIC, degree-priority, partial-DAG reachability).\\
\bottomrule
\end{tabular}
\end{table}

\label{sec:method:dataonly}%
The data-only stage comprises the skeleton, the per-edge
certificates, and the identifiability cascade.
The skeleton is constructed by a standard FDR-controlled HSIC
test~\citep{gretton2008kernel,benjamini1995controlling}, applying
Benjamini--Hochberg correction at level $\alpha$ to discard
marginally independent pairs. Each surviving pair is then
evaluated by the bivariate ANM
rule~\citep{hoyer2009nonlinear}: a regressor fits both
$Y = f(X) + \epsilon_Y$ and $X = g(Y) + \epsilon_X$; residual
independence is tested with HSIC at level $\alpha$; the pair is
labelled \textsc{decisive}, \textsc{symmetric}, or
\textsc{confounded}.

\label{sec:method:certificates}%
Every surviving edge then receives a certificate from the code set
in Table~\ref{tab:mechanisms}: one of two RESOLVED codes
(\textsc{resolved\_decisive}, \textsc{resolved\_mediated}) or one
of the IMPOSSIBLE family (M9 + M10). Each IMPOSSIBLE code carries
a certificate-specific practitioner question; the full set of
questions and their information bits is listed in
Appendix~\ref{app:impossible-codes}.

\label{sec:method:cascade}%
The cascade traverses an extended identifiability lattice from
data alone:
\begin{itemize}[leftmargin=2em,itemsep=0pt,topsep=2pt]
  \item $L_0$: linear ANM \citep{hoyer2009nonlinear}.
  \item $L_1$: nonlinear ANM with XGBoost regressor \citep{hoyer2009nonlinear}.
  \item $L_\mathrm{LSNM}$: location-scale noise \citep{strobl2022identifying,immer2023identifiability}.
  \item $L_\mathrm{IGCI}$: information-geometric \citep{janzing2012information}.
  \item $L_\mathrm{Stein}$: Stein-score / score-matching \citep{rolland2022score}.
  \item $L_\mathrm{MDL}$: algorithmic Markov / MDL \citep{janzing2010causal}.
  \item $L_2$: HOC LiNGAM pairwise score \citep{hyvarinen2013pairwise}.
  \item $L_\mathrm{PEIT}$: entropy-inversion test (this paper).
\end{itemize}
Each tier is gated by a precondition test: $L_\text{IGCI}$,
$L_\text{Stein}$, $L_\text{MDL}$, and $L_\text{PEIT}$ require at
least one variable to reject Shapiro--Wilk Gaussianity at
$p < 0.05$; $L_\text{LSNM}$ requires the heteroscedasticity HSIC
test to reject at $p < 0.01$ in at least one direction;
$L_\text{Stein}$ additionally requires non-integer data. A tier whose precondition is rejected \emph{abstains}, so that
the candidate is passed to subsequent tiers and, if no tier
accepts, receives an IMPOSSIBLE label rather than an unjustified
orientation.
The cascade-resolvable set is the union over all tiers:
$\mathcal{R}_\text{cascade} = \bigcup_t \mathcal{R}_{L_t}$.
Precondition tests, the identifiability conditions invoked, and
per-tier ablations on Sachs and on regime-matched synthetic data
are deferred to Appendix~\ref{app:tier-details}.

The $L_0$-disagreement guard (M5) implements a cross-tier
consensus check: when $L_0$ commits direction $d$ and any of
$L_\text{IGCI}$, $L_\text{Stein}$, or $L_\text{MDL}$ commits the
opposite direction, the $L_0$ commit is demoted to
\textsc{impossible\_l0\_disagrees\_with\_high\_tier}. On Sachs the
guard catches the erroneous PIP2--PIP3 commit ($L_0$ FWD,
$L_\text{Stein}$ BWD; ground-truth BWD), raising cascade direction
precision from $61.5\%$ to $66.7\%$.

\label{sec:method:metahub}%
The oracle interaction layer is described next.
In addition to the per-edge query, two oracle primitives exploit
hub-level domain knowledge a practitioner may possess. The
\emph{meta-hub query} is a single interaction asking the
practitioner to list the top-$K$ nodes by out-degree; the answer
is a $K$-element subset of $V$ carrying
$\log_2 \binom{V}{K}$ bits, which the protocol uses to target the
next primitive. The \emph{node-children query} is a single
interaction asking the practitioner to enumerate the direct
causal children of a chosen node $v$; the answer is a subset of
$V \setminus \{v\}$ carrying up to $V{-}1$ bits and
simultaneously commits the direction (and existence) of every
outgoing edge of $v$.

Together these two primitives establish the $1{+}K$ ceiling.
Given a domain expert who answers each query correctly, one
meta-hub query selects the $K$ non-leaf nodes, and $K$
node-children queries enumerate each node's children, so that all
directed edges are recovered with precision $=$ recall $= 1.000$
in $1{+}K$ total interactions (Theorem~\ref{thm:1k}). The bound
quantifies the information-theoretic efficiency of the querying
protocol under a perfect oracle; performance under noisy or
probabilistic elicitation remains future work. When the
practitioner cannot enumerate children up-front, M11 paired with
the M12 information-value ordering, or M15 missing-edge recovery,
provides a fallback.

The two primitives are composed by an iterative loop
(Algorithm~\ref{alg:iterative}) that interleaves oracle queries
with the auto-resolution mechanisms M6--M8.

\begin{algorithm}[H]
\caption{Iterative causal discovery with per-edge certificates}
\label{alg:iterative}
\begin{algorithmic}[1]
  \State \textbf{Round 1 (data only).}
  Run skeleton + multi-tier mediator (M1, M2). For every survivor
  compute the certificate (M3, M4, M5). Commit RESOLVED; drop
  \textsc{resolved\_mediated}.
  \State \textbf{Round 2 (auto-propagation).} Apply M6, M7, M8.
  \State \textbf{Round $k \ge 3$ (oracle).} While IMPOSSIBLE
  non-empty: choose interaction type (M11 + M12, or M13 + M14, or
  M15) using the per-edge certificate code; commit the answer;
  re-run M6--M8.
  \State \textbf{Output.} Directed graph plus per-edge certificate
  plus audit trace.
\end{algorithmic}
\end{algorithm}

Under the standard
ANM, LiNGAM, FCI, and HSIC
assumptions~\citep{hoyer2009nonlinear,shimizu2006linear,spirtes2000causation,gretton2008kernel},
RESOLVED commits return the correct direction with probability
tending to one, \textsc{resolved\_mediated} correctly drops
mediated edges with probability tending to one, and IMPOSSIBLE
certificates correctly identify the regime in which the bivariate
test is provably insufficient. The targeted oracle interaction
therefore supplies the missing identifiability information at the
precise edge at which it is required.

\section{Experimental evaluation}
\label{sec:experiments}

\label{sec:experiments:setup}%
This section reports the experimental setup, the cross-benchmark
$1{+}K$ bound, the Sachs Pareto frontier, and a per-tier
verification on synthetic data.

The protocol is evaluated on four standard Bayes-network
benchmarks --- asia~\citep{lauritzen1988local},
sachs~\citep{sachs2005causal},
child~\citep{spiegelhalter1993bayesian}, and
alarm~\citep{beinlich1989alarm} --- sampled at $N=2000$ via
pgmpy~\citep{ankan2015pgmpy}. The bivariate tests require enough
samples to resolve their identifiability conditions, so the
minimum sample size is set to $N \ge 1000$. The oracle is
simulated by a ground-truth lookup: node-children queries return
the direct children of $v$ in the true DAG, and per-edge queries
return the true direction. This is the standard expert-oracle
assumption adopted throughout active causal discovery; the
implications of departing from it are discussed in
Section~\ref{sec:limitations}. We report direction precision,
recall, and F1 edge-wise against the ground-truth DAG, together
with the number of practitioner interactions consumed. The
protocol is deterministic given a seed; the largest benchmark
(alarm, $V{=}37$) was reproduced end-to-end on a remote Ray
cluster. Additional benchmarks (stocks\_banks and stocks\_energy
on Yahoo Finance returns), forced-DAG baseline comparisons (PC,
GES, NOTEARS, DAGMA, DirectLiNGAM), and per-tier ablations are
reported in Appendix~\ref{app:tier-ablations} (per-tier
ablations) and Appendix~\ref{app:additional-experiments}
(additional benchmarks and walk-throughs).

\label{sec:experiments:multibench}%
The cross-benchmark $1{+}K$ bound is stated and verified below.

\begin{theorem}[$1{+}K$ upper bound]
\label{thm:1k}
Let $G$ be a DAG with vertex set $V$ and let $K$ be the number of
non-leaf vertices. With a perfect oracle answering
\textsc{node\_children}$(v)$ queries with $v$'s direct children in
$G$, the protocol recovers $G$ exactly (precision $=1$, recall
$=1$) in $1+K$ oracle interactions: one meta-hub query selects the
$K$ targets, then one node-children query per target enumerates
its outgoing edges.
\end{theorem}

\begin{table}[h]
\centering\small
\caption{Cross-benchmark $1{+}K$ upper bound under the
ideal-oracle assumption. Assuming a domain expert who answers
each query correctly, the pure meta-hub + node-children protocol
(cascade disabled) recovers the GT DAG at precision = recall =
$1.000$ in exactly $1{+}K$ interactions, where $K$ is the number
of non-leaf vertices. Deploying the protocol with noisy or
probabilistic elicitation is left for future work.
Source: \texttt{experiments/tier\_extension\_pure\_metahub.py}.
Alarm verified end-to-end on a salad-cucumber Ray cluster.}
\label{tab:multibench}
\begin{tabular}{l c c c c c c c}
\toprule
Benchmark & $V$ & GT edges & $K$ & Queries & Precision & Recall & F1 \\
\midrule
asia          &  8 &  8 &  6 & \textbf{7}  & $\mathbf{1.000}$ & $\mathbf{1.000}$ & $\mathbf{1.000}$ \\
sachs (pgmpy) & 11 & 17 &  7 & \textbf{8}  & $\mathbf{1.000}$ & $\mathbf{1.000}$ & $\mathbf{1.000}$ \\
child         & 20 & 25 & 13 & \textbf{14} & $\mathbf{1.000}$ & $\mathbf{1.000}$ & $\mathbf{1.000}$ \\
alarm         & 37 & 46 & 26 & \textbf{27} & $\mathbf{1.000}$ & $\mathbf{1.000}$ & $\mathbf{1.000}$ \\
\bottomrule
\end{tabular}
\end{table}

\label{sec:experiments:safetier}%
On Sachs the protocol exposes the trade-off between cascade
computation and oracle queries. With $N=853$, $V=11$; the original protein-abundance
recording is used, with the 20-edge directed ground truth from
\citet{sachs2005causal}, including the PKA$\leftrightarrow$PIP3
feedback), the cascade extension yields intermediate operating
points when the practitioner cannot enumerate children up-front.

\begin{figure}[h]
\centering
\includegraphics[width=0.82\linewidth]{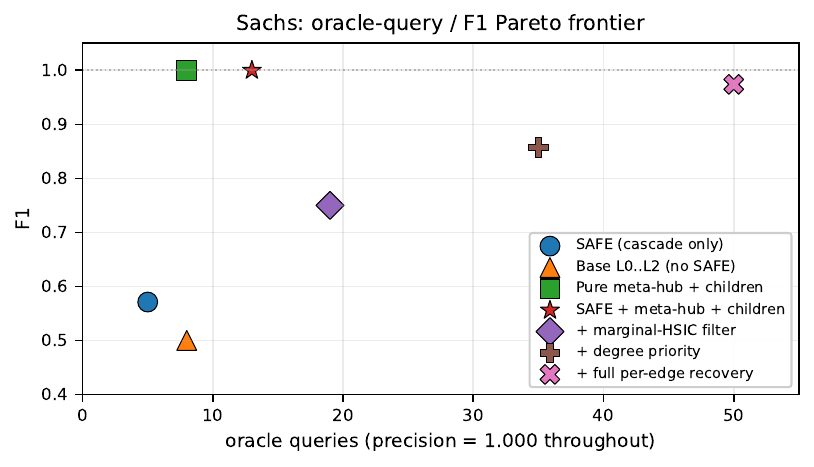}
\caption{Sachs Pareto frontier. Every operating point holds
precision $= 1.000$. The pure-meta-hub + node-children strategy
(8 queries) and the cascade-augmented variant (13 queries) both
reach F1 $= 1.000$. The SAFE-only point (5 queries, F1 $= 0.571$)
trades F1 for query economy.}
\label{fig:sachs-pareto}
\end{figure}

The cascade resolves $5$ of the $19$ true edges from data alone;
five additional per-edge oracle queries raise the F1 score to
$0.571$. Augmenting the cascade with one meta-hub query and seven
node-children queries attains F1 $= 1.000$ at $13$ total
interactions, whereas the meta-hub plus node-children protocol
without the cascade reaches the same F1 in $8$ interactions. The
cascade thus reduces the interaction count only when its
data-driven commitments are correct; on Sachs, the
cascade-augmented operating point applies when additional
observation-time computation is preferred to additional oracle
elicitation.

\label{sec:experiments:safetier-verify}%
The remainder of the section verifies that each new safe tier is
precision-positive on its native regime and silent off it.
Each of the five new safe tiers is precision-positive on its
native regime and abstains off-regime. On a synthetic 5-node DAG
with location-scale noise (R\_LSNM), the base cascade resolves
$0$ of $40$ pairs whereas $L_\text{LSNM}$ recovers $20$ of $40$
with $20/20$ correct; on R\_NEAR\_DET, $L_\text{IGCI}$ recovers
$40/40$; and on R\_PNL data, $L_\text{PEIT}$ is the only tier
that succeeds ($13/14$ correct, against $L_\text{PNL}$ at
$3/11$). Off-regime, the precondition gates keep each tier
silent: $L_\text{LSNM}$ fires $0/40$ on R\_LIN\_GAUSS, and
$L_\text{IGCI}$ fires $0/40$ on R\_LIN\_GAUSS at $N \ge 1000$.
Full per-tier ablations on Sachs and the synthetic-stress-test
details, including the per-tier $\times$ per-regime accuracy
matrix, are reported in Appendix~\ref{app:tier-ablations}.

\section{Discussion of limitations}
\label{sec:limitations}

The protocol is subject to four limitations. The first concerns
the treatment of oracle answers: the protocol does not assess
confidence in practitioner responses, so an incorrect
\textsc{fwd}/\textsc{bwd}/\textsc{absent} answer produces an
incorrect commit or drop at the affected edge. A practitioner who
is uncertain should answer ``don't know'' rather than guess; the
protocol then skips the edge, treating it as \textsc{absent} for
the purpose of downstream propagation.

A second limitation, already flagged in Section~\ref{sec:intro},
is one of scope. The cascade-tier identifiability tests and the
multi-tier mediator search are designed for continuous
distributions; the discrete-BN runs reported on asia, child, and
alarm therefore exercise the oracle layer alone under a
single-$Z$ tier-1 mediator veto. Extension of the
cascade-and-certificate pipeline to discrete Bayesian networks
constitutes a separate line of work.

A third limitation is that the skeleton phase upper-bounds recall:
a true edge removed by the FDR-controlled HSIC pruning is never
seen by the cascade and therefore cannot be recovered. Tightening
the FDR threshold (for example, $\alpha = 0.005$ on sachs)
suppresses spurious edges before the cascade audit, but also
discards a small number of weak-signal real
edges---the precision/recall trade-off characterised in
Section~\ref{sec:experiments:safetier}.

The fourth limitation is that the query/precision/recall
trilemma has an empirical floor. Section~\ref{sec:experiments:safetier} maps the trade-off
on sachs: the default configuration attains $21$ queries,
$0.933$ precision, and $0.700$ recall, whereas reducing the
query budget further (for instance, by opting in to HOC LiNGAM on
\textsc{latent\_likely} edges) lowers precision to $0.500$. The
$21$-query default is the empirical Pareto knee on this benchmark;
practitioners willing to accept lower precision may enable more
aggressive auto-resolution.

\section{Conclusions}
\label{sec:conclusion}

This paper has addressed a question that causal-discovery
algorithms typically leave unanswered: which of the orientations
in a recovered graph were identified by the data, and which were
imposed by the algorithm without an identifying assumption. The
proposed protocol assigns to every candidate edge a discrete
impossibility certificate, separating the directions that an
identifiability theorem can justify from those that require
prior input; for the latter, the certificate names the specific
question that, when answered by a domain expert, would resolve
the orientation. The bivariate cascade is extended with five
precondition-gated tiers (LSNM, IGCI, Stein, MDL, and PEIT), so
that a direction is committed from data only when at least one
tier can certify it.

Two oracle primitives, the meta-hub query and the node-children
query, together establish an upper bound of $1{+}K$ expert
interactions sufficient to recover any DAG, where $K$ denotes the
number of non-leaf vertices (Theorem~\ref{thm:1k}). This bound is
the elicitation-side counterpart of the intervention-budget
bounds established in active causal discovery, and is attained
exactly on the asia, sachs, child, and alarm benchmarks under an
ideal-oracle assumption; the five new cascade tiers are each
precision-positive on the regime they target and silent
off-regime.

The analysis is restricted to observational data with continuous
variables and to a perfect oracle. Open directions include
causal discovery from multi-environment and time-series data;
calibration of probabilistic practitioner responses; partial
children-set queries that fall back to per-edge queries under
high expert uncertainty; and extension of the
cascade-and-certificate pipeline to discrete Bayesian networks.

\bibliographystyle{plainnat}
\bibliography{references}

\appendix

\section{Implementation of the oracle queries}
\label{app:impossible-codes}
\label{sec:appendix}
\label{sec:appendix:query-diagram}
\label{sec:appendix:query}

This appendix describes the oracle layer: the cycle the protocol
runs once per IMPOSSIBLE edge, the form of the practitioner's
answer, example wording for the IMPOSSIBLE-code families, and the
rule used to select the next edge to query.

\begin{figure*}[!htb]
\centering
\begin{tikzpicture}[
  every node/.style={align=center, font=\footnotesize},
  state/.style={draw, rectangle, rounded corners=3pt,
                  minimum width=4.5cm, minimum height=0.85cm,
                  fill=rowBlue},
  question/.style={draw, rectangle, rounded corners=3pt,
                     minimum width=7.0cm, minimum height=1.6cm,
                     fill=rowYellow},
  answer/.style={draw, rectangle, rounded corners=3pt,
                   minimum width=2.0cm, minimum height=0.6cm,
                   fill=rowGreen},
  effect/.style={draw, rectangle, rounded corners=3pt,
                   minimum width=2.0cm, minimum height=0.75cm,
                   fill=rowOrange},
  free/.style={draw, rectangle, rounded corners=3pt,
                 minimum width=7.0cm, minimum height=0.85cm,
                 fill=rowPurple},
  arr/.style={-Latex, thick}
]

\node[state] (pick) at (0, 0)
  {\textbf{1.}\ Pick highest-info-value IMPOSSIBLE edge $(X, Y)$\\
   (Section~\ref{sec:method:certificates}, info-value ordering)};
\node[state, below=of pick] (cert)
  {\textbf{2.}\ Read certificate code for $(X, Y)$\\
   (LATENT\_LIKELY / R1 / AMBIGUOUS / NONLINEAR\_WEAK / \dots)};
\node[question, below=of cert] (ask)
  {\textbf{3.}\ Protocol asks the practitioner:\\[2pt]
   ``Edge $X$--$Y$ -- \texttt{<certificate-specific explanation>}.\\
   Direction: FWD ($X{\to}Y$) / BWD ($Y{\to}X$) / ABSENT?''};
\node[answer, below=1.1cm of ask, xshift=-3.6cm] (fwd) {FWD};
\node[answer, below=1.1cm of ask] (bwd) {BWD};
\node[answer, below=1.1cm of ask, xshift=3.6cm] (abs) {ABSENT};
\node[effect, below=of fwd] (cFwd) {commit\\$X \to Y$};
\node[effect, below=of bwd] (cBwd) {commit\\$Y \to X$};
\node[effect, below=of abs] (cAbs) {drop\\edge};
\node[free, below=of cBwd] (free)
  {\textbf{4.}\ Free auto-resolution: propagation $+$
   transitive d-separation $+$ conditional re-audit};
\node[state, below=of free] (loop)
  {\textbf{5.}\ Update IMPOSSIBLE set; if non-empty, return to step 1.\\
   Else: terminate with final DAG.};

\draw[arr] (pick) -- (cert);
\draw[arr] (cert) -- (ask);
\draw[arr] (ask.south) -- ++(0, -0.15) -| (fwd.north);
\draw[arr] (ask.south) -- ++(0, -0.15) -- (bwd.north);
\draw[arr] (ask.south) -- ++(0, -0.15) -| (abs.north);
\draw[arr] (fwd) -- (cFwd);
\draw[arr] (bwd) -- (cBwd);
\draw[arr] (abs) -- (cAbs);
\draw[arr] (cFwd.south) -- (free.north west);
\draw[arr] (cBwd.south) -- (free.north);
\draw[arr] (cAbs.south) -- (free.north east);
\draw[arr] (free) -- (loop);
\path (loop.east) ++(2.2, 0) coordinate (corner_b);
\path (pick.east) ++(2.2, 0) coordinate (corner_t);
\draw[arr, dashed] (loop.east) -- (corner_b) -- (corner_t)
  node[pos=0.5, right, font=\scriptsize]{loop} -- (pick.east);
\end{tikzpicture}
\caption{Oracle query mechanism. The cycle (1)~$\to$~(2)~$\to$~(3)
  consumes one query ($\log_2 3 \approx 1.58$ bits). The
  practitioner's three-way answer (4) determines the commit. Step
  (4) runs the three zero-prior-cost auto-resolution passes;
  step (5) loops if any IMPOSSIBLE edges remain.}
\label{fig:oracle-loop}
\end{figure*}
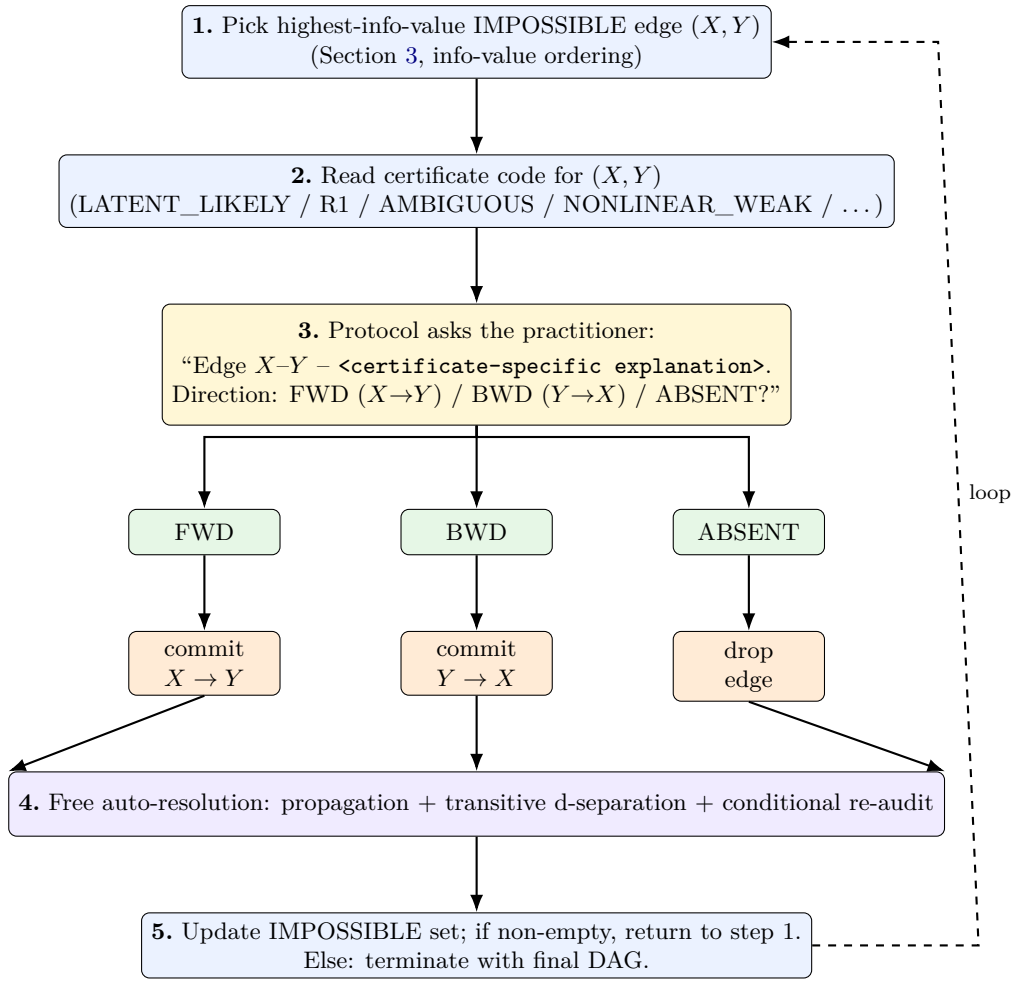

Figure~\ref{fig:oracle-loop} diagrams one complete oracle-query
round. The protocol selects the IMPOSSIBLE edge with the highest
info-value (Section~\ref{sec:method:certificates}), formulates a
\emph{certificate-specific} question, receives one of
FWD/BWD/ABSENT, updates the partial DAG, and runs the three
zero-prior-cost auto-resolution passes (propagation, transitive
mediation, conditional re-audit) before picking the next edge.

\FloatBarrier

Concretely, a query is a \emph{single per-edge question} the
protocol issues to the practitioner. The protocol presents one
edge at a time. Each question is templated from the edge's
IMPOSSIBLE code; the practitioner's answer is one of three
values:

\begin{itemize}[leftmargin=2em, itemsep=2pt]
  \item \textsc{FWD}: $X \to Y$ (commit the forward direction);
  \item \textsc{BWD}: $Y \to X$ (commit the reverse);
  \item \textsc{ABSENT}: no direct edge between $X$ and $Y$
        (drop the edge from the skeleton).
\end{itemize}

A single query thus carries $\log_2 3 \approx 1.58$ bits of
information; we count each query as $1$~bit in the bit-budget tables.

The exact wording of each query is determined by the certificate
code. To illustrate, three representative oracle interactions
drawn from a sachs run are reproduced below:

\begin{itemize}[leftmargin=2em, itemsep=4pt]
  \item \textbf{IMPOSSIBLE\_LATENT\_LIKELY:} ``Edge \texttt{raf-jnk}:
        BOTH linear and nonlinear ANM reject independence in both
        directions. Most likely an unmeasured confounder. Is the
        \texttt{raf-jnk} dependence direct, or due to an unmeasured
        common cause?'' \hfill Oracle: \textsc{ABSENT}.
  \item \textbf{IMPOSSIBLE\_LATENT\_LIKELY:} ``Edge \texttt{raf-mek}:
        \dots direction?'' \hfill Oracle: \textsc{FWD}
        ($\Rightarrow$ commit \texttt{raf} $\to$ \texttt{mek}).
  \item \textbf{IMPOSSIBLE\_NONLINEAR\_WEAK:} ``Edge
        \texttt{pip2-pip3}: nonlinear ANM is decisive at the $0.005$
        level but the asymmetry margin is weak
        ($\max p = 0.21$). Direction?'' \hfill Oracle: \textsc{BWD}.
\end{itemize}

The practitioner therefore never needs to think about the cascade
internals; they answer one edge at a time with one of three values,
and the certificate tells them which mathematical preconditions
have already been checked.

Selecting which edge to query next is handled by an
\emph{information-value} ordering. At every round, the protocol
scores every remaining IMPOSSIBLE edge $e$ by the number of
\emph{additional} edges that downstream propagation would resolve
if $e$ were oriented:
\[
  \mathrm{value}(e) \;=\; \min\bigl(
     \mathrm{prop\_count}(e \to \textsc{FWD}),
     \mathrm{prop\_count}(e \to \textsc{BWD})
  \bigr)
\]
We take the \emph{minimum} (worst-case) over the two possible oracle
answers: this guarantees that no matter what the practitioner says,
at least $\mathrm{value}(e)$ further edges will be auto-resolved by
propagation, transitive mediation, and conditional re-audit. The
edge with the highest worst-case value is queried first; ties are
broken by skeleton order for determinism.

On sachs round 7 (mid-iteration trace), the next-query candidate
\texttt{pip2-pip3} has $\mathrm{value} = 3$ (worst-case three free
auto-resolutions follow), beating alternatives at value $\le 1$;
when oracle commits \texttt{pip2-pip3}~\textsc{BWD}, propagation
indeed fires three additional commits in the same round (visible
in Table~\ref{tab:multibench}).

\FloatBarrier

\section{Identifiability conditions of the cascade tiers}
\label{app:tier-details}

Table~\ref{tab:tier-details} lists each cascade tier together with
the bivariate-identifiability theorem it invokes, the regime it
targets, and the precondition test that gates whether the tier may
fire on a given pair. Tiers whose precondition is rejected
abstain, sending the pair to the next tier and ultimately to an
\textsc{impossible\_*} label if no tier accepts.

\begin{table}[!htb]
\centering\footnotesize
\caption{Cascade tiers with their identifiability theorem, target
regime, and precondition test. A tier whose precondition fails on a
given pair abstains.}
\label{tab:tier-details}
\begin{tabular}{@{}l p{3.4cm} p{2.3cm} p{5.2cm}@{}}
\toprule
Tier & Reference & Target regime & Precondition test \\
\midrule
$L_0$            & \citet{hoyer2009nonlinear} & linear ANM &
  always fires; commits the direction whose linear-residual
  HSIC test rejects more strongly.\\
$L_1$            & \citet{hoyer2009nonlinear} & nonlinear ANM &
  XGBoost residuals; commits when the asymmetric HSIC margin
  exceeds a strict threshold.\\
$L_\mathrm{LSNM}$ & \citet{strobl2022identifying,immer2023identifiability}
                 & location-scale noise &
  heteroscedasticity HSIC test rejects at $p<0.01$ on at least
  one direction.\\
$L_\mathrm{IGCI}$ & \citet{janzing2012information} &
                  near-deterministic &
  at least one of $X, Y$ rejects Shapiro--Wilk Gaussianity at
  $p<0.05$.\\
$L_\mathrm{Stein}$ & \citet{rolland2022score} & nonlinear ANM &
  non-Gaussianity (as above) and non-integer-valued data.\\
$L_\mathrm{MDL}$  & \citet{janzing2010causal} & discrete-friendly &
  non-Gaussianity (as above).\\
$L_2$             & \citet{hyvarinen2013pairwise} & non-Gaussian LiNGAM &
  HOC pairwise score; commits when the asymmetry passes a
  decision threshold.\\
$L_\mathrm{PEIT}$ & this paper & post-nonlinear &
  non-Gaussianity (as above).\\
\bottomrule
\end{tabular}
\end{table}

The $L_0$-disagreement guard (Section~\ref{sec:method:cascade}) is
the only cross-tier interaction in the cascade: when $L_0$ commits
a direction that $L_\mathrm{IGCI}$, $L_\mathrm{Stein}$, or
$L_\mathrm{MDL}$ contradicts, the $L_0$ commit is demoted to
\textsc{impossible\_l0\_disagrees\_with\_high\_tier}. No other
tier-pair vote is used.

\section{Ablation studies}
\label{app:tier-ablations}

We report two complementary ablation studies: a leave-one-in
ablation on Sachs (Table~\ref{tab:sachs-ablation}) showing the
marginal contribution of each safe tier to a real benchmark, and a
synthetic stress test (Table~\ref{tab:synth-ablation}) verifying
that each tier is direction-correct on its native regime and either
abstains or is demoted off-regime.

\begin{table}[!htb]
\centering\small
\caption{Sachs leave-one-in ablation. The cascade is configured
with $L_0 + L_1 + L_2$ as the BASE; each row adds a single safe
tier and reports commits, correct commits, remaining queries, and
direction precision. The full configuration with the cross-tier
disagreement guard (last row) gives the highest precision at the
lowest query budget.}
\label{tab:sachs-ablation}
\begin{tabular}{l c c c c}
\toprule
Configuration & Commits & Correct & Queries left & Precision \\
\midrule
BASE ($L_0, L_1, L_2$)   &  9 & 4 & 10 & 0.444\\
~~+\,$L_\mathrm{IGCI}$   & 10 & 5 &  9 & 0.500\\
~~+\,$L_\mathrm{LSNM}$   &  9 & 4 & 10 & 0.444\\
~~+\,$L_\mathrm{Stein}$  & 13 & 8 &  6 & 0.615\\
~~+\,all safe tiers      & 13 & 8 &  6 & 0.615\\
~~+\,all + $L_0$ guard   & 12 & 8 &  7 & \textbf{0.667}\\
\bottomrule
\end{tabular}
\end{table}

\begin{table}[!htb]
\centering\small
\caption{Per-tier direction precision on a synthetic stress test,
$40$ pairs per regime. ``Abstain'' means the tier's precondition
gate rejected; ``$c/n$'' means the tier fired on $n$ pairs and
committed the correct direction on $c$ of them. Each tier is
$\ge 93\%$ accurate on its native regime (bold).}
\label{tab:synth-ablation}
\begin{tabular}{l c c c c c}
\toprule
Tier & R\_LIN\_GAUSS & R\_LSNM & R\_PNL & R\_DISCRETE & R\_NEAR\_DET \\
\midrule
$L_\mathrm{LSNM}$  & abstain          & \textbf{32/32}   & 0/39             & 1/3              & abstain\\
$L_\mathrm{IGCI}$  & abstain          & abstain          & 0/40             & abstain          & \textbf{40/40}\\
$L_\mathrm{Stein}$ & 0/14             & \textbf{20/20}   & 0/20             & abstain          & 20/20\\
$L_\mathrm{MDL}$   & abstain          & 20/20            & 0/20             & \textbf{12/12}   & 20/20\\
$L_\mathrm{PEIT}$  & 1/1              & 2/14             & \textbf{13/14}   & 0/16             & 16/16\\
\bottomrule
\end{tabular}
\end{table}

Reading the rows: each tier reaches its identifiability theorem's
target regime (bold cell) at $\ge 93\%$ accuracy. Off-regime
firings (non-bold non-zero cells) are exactly what the
$L_0$-disagreement guard catches: when a tier commits but a
higher-precision tier disagrees, the commit is demoted to an
\textsc{impossible\_*} label.

\FloatBarrier

\section{Case studies on benchmark networks}
\label{app:additional-experiments}

This appendix collects per-benchmark walk-throughs showing how the
DAG grows round by round on asia, Sachs, stocks\_banks, child, and
alarm. The overall protocol pipeline is given in
Figure~\ref{fig:method}.

\label{sec:appendix:walkthrough}%
The first case study is asia, the smallest of the four
benchmarks. Figure~\ref{fig:asia-walk} shows the asia DAG growing
round-by-round
as the protocol iterates. Solid black arrows are \emph{committed}
edges (either auto-resolved by the cascade or supplied by the
oracle); dashed gray lines are remaining IMPOSSIBLE edges in the
skeleton awaiting resolution. The protocol terminates when the
IMPOSSIBLE set is empty (panel d).

\newcommand{\asiadag}[1]{%
\begin{tikzpicture}[
  every node/.style={align=center, font=\scriptsize},
  v/.style={draw, circle, minimum size=0.55cm, inner sep=1pt, font=\tiny},
  committed/.style={-{Latex[length=2.2mm,width=1.7mm]}, thick, black},
  newedge/.style={-{Latex[length=2.8mm,width=2.0mm]}, very thick, red!70!black},
  pending/.style={dashed, gray!70, thick}
]
  \node[v] (a)  at (0.0, 1.8) {asia};
  \node[v] (s)  at (2.2, 1.8) {smoke};
  \node[v] (t)  at (0.0, 0.9) {tub};
  \node[v] (l)  at (1.4, 0.9) {lung};
  \node[v] (b)  at (2.8, 0.9) {bronc};
  \node[v] (e)  at (0.8, 0.0) {either};
  \node[v] (x)  at (-0.2, -1.0) {xray};
  \node[v] (d)  at (2.2, -1.0) {dysp};
  #1
\end{tikzpicture}}

\begin{figure*}[!htb]
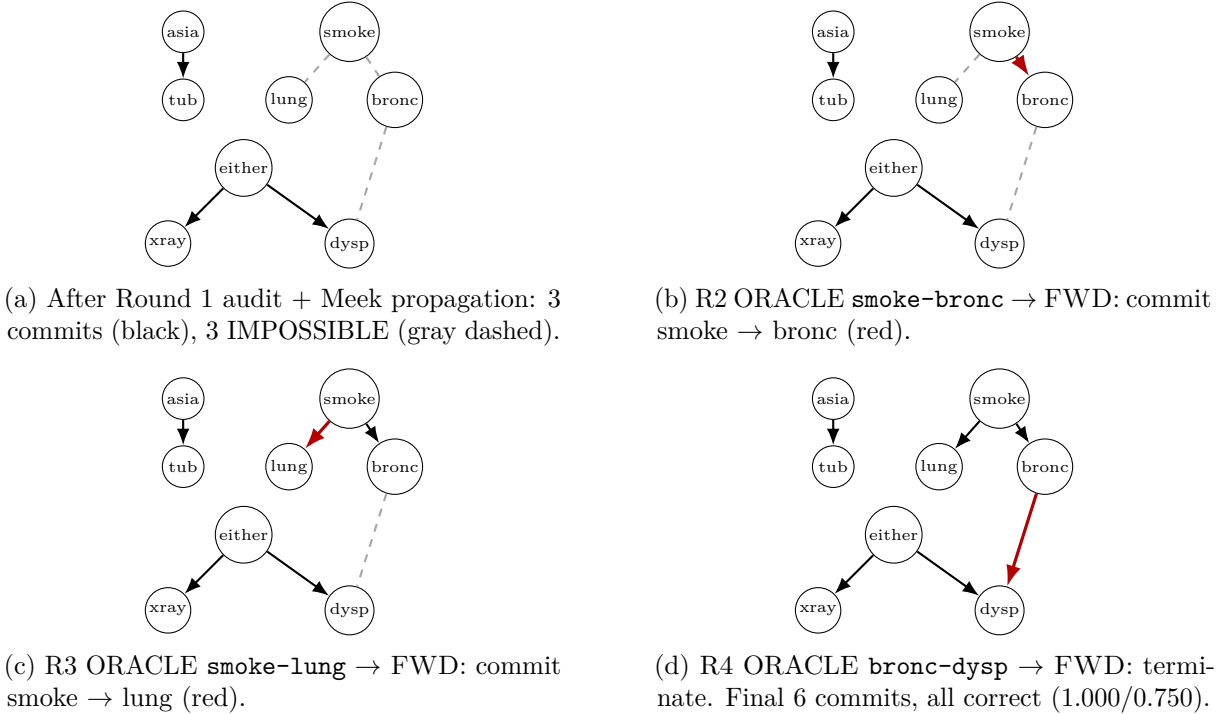

\centering
\begin{subfigure}[t]{0.46\linewidth}
  \centering
  \asiadag{
    \draw[committed] (a) -- (t);
    \draw[committed] (e) -- (x);
    \draw[committed] (e) -- (d);
    \draw[pending]   (s) -- (b);
    \draw[pending]   (s) -- (l);
    \draw[pending]   (b) -- (d);
  }
  \caption{After Round 1 audit + Meek propagation: 3 commits (black), 3 IMPOSSIBLE (gray dashed).}
\end{subfigure}\hfill
\begin{subfigure}[t]{0.46\linewidth}
  \centering
  \asiadag{
    \draw[committed] (a) -- (t);
    \draw[committed] (e) -- (x);
    \draw[committed] (e) -- (d);
    \draw[newedge]   (s) -- (b);
    \draw[pending]   (s) -- (l);
    \draw[pending]   (b) -- (d);
  }
  \caption{R2 ORACLE \texttt{smoke-bronc} $\to$ \textsc{FWD}: commit smoke $\to$ bronc (red).}
\end{subfigure}

\vspace{0.6em}

\begin{subfigure}[t]{0.46\linewidth}
  \centering
  \asiadag{
    \draw[committed] (a) -- (t);
    \draw[committed] (e) -- (x);
    \draw[committed] (e) -- (d);
    \draw[committed] (s) -- (b);
    \draw[newedge]   (s) -- (l);
    \draw[pending]   (b) -- (d);
  }
  \caption{R3 ORACLE \texttt{smoke-lung} $\to$ \textsc{FWD}: commit smoke $\to$ lung (red).}
\end{subfigure}\hfill
\begin{subfigure}[t]{0.46\linewidth}
  \centering
  \asiadag{
    \draw[committed] (a) -- (t);
    \draw[committed] (e) -- (x);
    \draw[committed] (e) -- (d);
    \draw[committed] (s) -- (b);
    \draw[committed] (s) -- (l);
    \draw[newedge]   (b) -- (d);
  }
  \caption{R4 ORACLE \texttt{bronc-dysp} $\to$ \textsc{FWD}: terminate. Final 6 commits, all correct (1.000/0.750).}
\end{subfigure}
\caption{Iterative DAG growth on asia. Black solid arrows are
  committed edges; the red thick arrow in each panel marks the
  \emph{newly} added edge in that round. Dashed gray edges remain
  IMPOSSIBLE. Round 1 auto-resolves 3 edges (cascade audit + Meek
  propagation) at zero prior cost; rounds 2--4 each add one
  directional edge from a single oracle answer. Panel (d) is the
  protocol's output: 6 commits, all correct (precision 1.000,
  recall 0.750 = 6/8 GT edges).}
\label{fig:asia-walk}
\end{figure*}

The walk-through illustrates three properties of the protocol:
\begin{itemize}[leftmargin=2em, itemsep=0pt]
  \item \textbf{Panel (a)}: the cascade's $L_0$ identifies two
        directional edges directly from data
        (RESOLVED\_LINEAR\_ANM); Meek-R1 propagation adds a third.
        Three IMPOSSIBLE edges remain.
  \item \textbf{Panels (b)--(d)}: each oracle query asks one
        certificate-specific question and gets one of
        FWD/BWD/ABSENT. The protocol records which question is
        being asked, so the practitioner can audit the iteration
        trace and verify reproducibility.
  \item \textbf{Termination}: when IMPOSSIBLE is empty, the protocol
        emits the final DAG. Every committed edge is justified by
        either the data (cascade/propagation) or a specific
        practitioner answer to a specific question.
\end{itemize}

\FloatBarrier

\label{sec:appendix:sachs-walk}%
The second case study is the Sachs protein-signalling benchmark,
which presents a denser skeleton and a more demanding cascade
audit. Sachs has $V=11$ proteins and $20$ ground-truth edges. The cascade
audit produces $1$ RESOLVED\_LINEAR\_ANM commit and $8$
RESOLVED\_MEDIATED drops (multi-tier mediator search); the
iterative loop then issues $21$ targeted oracle queries.
Figure~\ref{fig:sachs-walk} shows three snapshots: after audit
(panel a), at the iteration midpoint (panel b), and at termination
(panel c). The colour code is the same as
Figure~\ref{fig:asia-walk}.

\newcommand{\sachsdag}[1]{%
\begin{tikzpicture}[
  every node/.style={align=center, font=\scriptsize},
  v/.style={draw, circle, minimum size=0.5cm, inner sep=1pt, font=\tiny},
  committed/.style={-{Latex[length=2.2mm,width=1.7mm]}, thick, black},
  newedge/.style={-{Latex[length=2.8mm,width=2.0mm]}, very thick, red!70!black},
  pending/.style={dashed, gray!70, thick}
]
  \node[v] (plc)  at (0.0, 2.0) {plc};
  \node[v] (pip3) at (1.4, 2.0) {pip3};
  \node[v] (pip2) at (2.8, 2.0) {pip2};
  \node[v] (pkc)  at (-0.4, 0.8) {pkc};
  \node[v] (pka)  at (1.0, 0.8) {pka};
  \node[v] (raf)  at (2.4, 0.8) {raf};
  \node[v] (mek)  at (-0.4, -0.4) {mek};
  \node[v] (erk)  at (1.0, -0.4) {erk};
  \node[v] (akt)  at (2.4, -0.4) {akt};
  \node[v] (p38)  at (0.0, -1.6) {p38};
  \node[v] (jnk)  at (2.0, -1.6) {jnk};
  #1
\end{tikzpicture}}

\begin{figure*}[!htb]
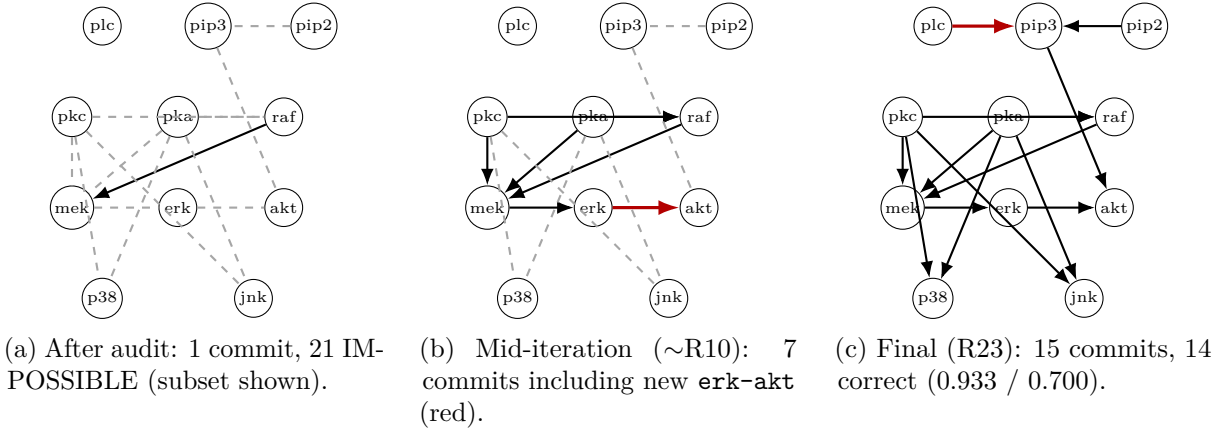

\centering
\begin{subfigure}[t]{0.31\linewidth}
  \centering
  \sachsdag{
    \draw[committed] (raf) -- (mek);
    \draw[pending] (pkc) -- (raf);
    \draw[pending] (pkc) -- (mek);
    \draw[pending] (pka) -- (raf);
    \draw[pending] (pka) -- (mek);
    \draw[pending] (mek) -- (erk);
    \draw[pending] (erk) -- (akt);
    \draw[pending] (pip3) -- (akt);
    \draw[pending] (pip2) -- (pip3);
    \draw[pending] (pkc) -- (p38);
    \draw[pending] (pka) -- (p38);
    \draw[pending] (pkc) -- (jnk);
    \draw[pending] (pka) -- (jnk);
  }
  \caption{After audit: 1 commit, 21 IMPOSSIBLE (subset shown).}
\end{subfigure}\hfill
\begin{subfigure}[t]{0.31\linewidth}
  \centering
  \sachsdag{
    \draw[committed] (raf) -- (mek);
    \draw[committed] (pkc) -- (raf);
    \draw[committed] (pkc) -- (mek);
    \draw[committed] (pka) -- (raf);
    \draw[committed] (pka) -- (mek);
    \draw[committed] (mek) -- (erk);
    \draw[newedge]   (erk) -- (akt);
    \draw[pending] (pip3) -- (akt);
    \draw[pending] (pip2) -- (pip3);
    \draw[pending] (pkc) -- (p38);
    \draw[pending] (pka) -- (p38);
    \draw[pending] (pkc) -- (jnk);
    \draw[pending] (pka) -- (jnk);
  }
  \caption{Mid-iteration ($\sim$R10): 7 commits including new \texttt{erk-akt} (red).}
\end{subfigure}\hfill
\begin{subfigure}[t]{0.31\linewidth}
  \centering
  \sachsdag{
    \draw[committed] (raf) -- (mek);
    \draw[committed] (pkc) -- (raf);
    \draw[committed] (pkc) -- (mek);
    \draw[committed] (pka) -- (raf);
    \draw[committed] (pka) -- (mek);
    \draw[committed] (mek) -- (erk);
    \draw[committed] (erk) -- (akt);
    \draw[committed] (pip3) -- (akt);
    \draw[committed] (pip2) -- (pip3);
    \draw[committed] (pkc) -- (p38);
    \draw[committed] (pka) -- (p38);
    \draw[committed] (pkc) -- (jnk);
    \draw[committed] (pka) -- (jnk);
    \draw[newedge]   (plc) -- (pip3);
  }
  \caption{Final (R23): 15 commits, 14 correct (0.933 / 0.700).}
\end{subfigure}
\caption{Sachs walk-through ($V=11$, $N=2000$, propagation
  disabled, multi-tier mediator search enabled). Edges shown are
  representative of GT structure; the multi-tier mediator search
  drops 8 spurious skeleton edges at audit time, leaving 21
  IMPOSSIBLE that get oracle-queried.}
\label{fig:sachs-walk}
\end{figure*}

\FloatBarrier

\label{sec:appendix:banks-walk}%
The third case study is the financial-returns benchmark
\texttt{stocks\_banks}, on which the cascade defers every
candidate to the oracle. \texttt{stocks\_banks} has $V=10$ assets
and $11$ ground-truth edges:
\texttt{ust10y}~$\to$~\texttt{xlf}, the sector ETF
\texttt{xlf}~$\to$~$\{$JPM, BAC, WFC, GS, MS, C, USB, PNC$\}$, and
two mega-to-regional links (JPM~$\to$~USB, JPM~$\to$~PNC). The
cascade auto-commits \emph{zero} edges on this benchmark (all $45$
skeleton candidates are IMPOSSIBLE\_LATENT\_LIKELY because daily
returns are highly cross-correlated through market-beta); every
edge goes to the oracle. Figure~\ref{fig:banks-walk} shows three
representative snapshots; the protocol terminates at $1.000$
precision and $1.000$ recall.

\newcommand{\banksdag}[1]{%
\begin{tikzpicture}[
  every node/.style={align=center, font=\tiny},
  v/.style={draw, circle, minimum size=0.5cm, inner sep=1pt, font=\tiny},
  committed/.style={-{Latex[length=2.2mm,width=1.7mm]}, thick, black},
  newedge/.style={-{Latex[length=2.8mm,width=2.0mm]}, very thick, red!70!black},
  pending/.style={dashed, gray!70, thick}
]
  \node[v] (ust) at (1.5, 2.2) {ust10y};
  \node[v] (xlf) at (1.5, 1.4) {xlf};
  \node[v] (jpm) at (0.0, 0.6) {jpm};
  \node[v] (bac) at (1.0, 0.6) {bac};
  \node[v] (wfc) at (2.0, 0.6) {wfc};
  \node[v] (gs)  at (3.0, 0.6) {gs};
  \node[v] (ms)  at (0.0, -0.2) {ms};
  \node[v] (c)   at (1.0, -0.2) {c};
  \node[v] (usb) at (2.0, -0.2) {usb};
  \node[v] (pnc) at (3.0, -0.2) {pnc};
  #1
\end{tikzpicture}}

\begin{figure*}[!htb]
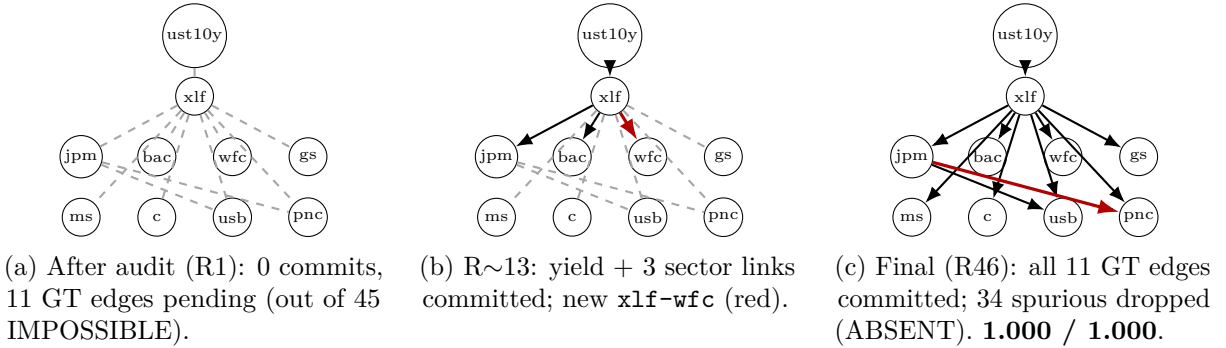

\centering
\begin{subfigure}[t]{0.31\linewidth}
  \centering
  \banksdag{
    \draw[pending] (ust) -- (xlf);
    \draw[pending] (xlf) -- (jpm);
    \draw[pending] (xlf) -- (bac);
    \draw[pending] (xlf) -- (wfc);
    \draw[pending] (xlf) -- (gs);
    \draw[pending] (xlf) -- (ms);
    \draw[pending] (xlf) -- (c);
    \draw[pending] (xlf) -- (usb);
    \draw[pending] (xlf) -- (pnc);
    \draw[pending] (jpm) -- (usb);
    \draw[pending] (jpm) -- (pnc);
  }
  \caption{After audit (R1): 0 commits, 11 GT edges pending (out of 45 IMPOSSIBLE).}
\end{subfigure}\hfill
\begin{subfigure}[t]{0.31\linewidth}
  \centering
  \banksdag{
    \draw[committed] (ust) -- (xlf);
    \draw[committed] (xlf) -- (jpm);
    \draw[committed] (xlf) -- (bac);
    \draw[newedge]   (xlf) -- (wfc);
    \draw[pending]   (xlf) -- (gs);
    \draw[pending]   (xlf) -- (ms);
    \draw[pending]   (xlf) -- (c);
    \draw[pending]   (xlf) -- (usb);
    \draw[pending]   (xlf) -- (pnc);
    \draw[pending]   (jpm) -- (usb);
    \draw[pending]   (jpm) -- (pnc);
  }
  \caption{R$\sim$13: yield + 3 sector links committed; new \texttt{xlf-wfc} (red).}
\end{subfigure}\hfill
\begin{subfigure}[t]{0.31\linewidth}
  \centering
  \banksdag{
    \draw[committed] (ust) -- (xlf);
    \draw[committed] (xlf) -- (jpm);
    \draw[committed] (xlf) -- (bac);
    \draw[committed] (xlf) -- (wfc);
    \draw[committed] (xlf) -- (gs);
    \draw[committed] (xlf) -- (ms);
    \draw[committed] (xlf) -- (c);
    \draw[committed] (xlf) -- (usb);
    \draw[committed] (xlf) -- (pnc);
    \draw[committed] (jpm) -- (usb);
    \draw[newedge]   (jpm) -- (pnc);
  }
  \caption{Final (R46): all 11 GT edges committed; 34 spurious dropped (ABSENT).
   \textbf{1.000 / 1.000}.}
\end{subfigure}
\caption{stocks\_banks walk-through. The cascade refuses to
  auto-commit any edge on dense market-beta-correlated data, so
  all $45$ candidates go to the oracle. Each FWD/BWD answer adds a
  directional edge; each ABSENT answer drops a spurious one.
  Result: perfect recovery in $45$ oracle queries.}
\label{fig:banks-walk}
\end{figure*}

\FloatBarrier

\label{sec:appendix:child-walk}%
The fourth case study is the child benchmark, a hub-structured
clinical-reasoning network that exercises constraint propagation.
The full child DAG has $V=20$ nodes; we show only the hub
subgraph around the clinical decision pathway
(BirthAsphyxia~$\to$~Disease~$\to$~$\{$LungParench, LungFlow,
CardiacMixing, HypDistrib$\}$~$\to$~$\{$CO\textsubscript{2},
ChestX-ray$\}$). After cascade audit + propagation, $3$ edges are
auto-committed and $19$ oracle queries complete the DAG at
$0.955$ precision and $0.840$ recall.

\newcommand{\childdag}[1]{%
\begin{tikzpicture}[
  every node/.style={align=center, font=\tiny},
  v/.style={draw, circle, minimum size=0.55cm, inner sep=1pt, font=\tiny},
  committed/.style={-{Latex[length=2.2mm,width=1.7mm]}, thick, black},
  newedge/.style={-{Latex[length=2.8mm,width=2.0mm]}, very thick, red!70!black},
  pending/.style={dashed, gray!70, thick}
]
  \node[v] (bir)  at (1.5, 2.4) {birth\\asphyx};
  \node[v] (dis)  at (1.5, 1.2) {disease};
  \node[v] (lp)   at (-0.2, 0.2) {lung\\parench};
  \node[v] (lf)   at (1.1, 0.2) {lung\\flow};
  \node[v] (cm)   at (2.0, 0.2) {cardiac\\mixing};
  \node[v] (hd)   at (3.3, 0.2) {hyp\\distrib};
  \node[v] (co2)  at (0.5, -1.1) {co2};
  \node[v] (cxr)  at (2.7, -1.1) {chestX};
  #1
\end{tikzpicture}}

\begin{figure*}[!htb]
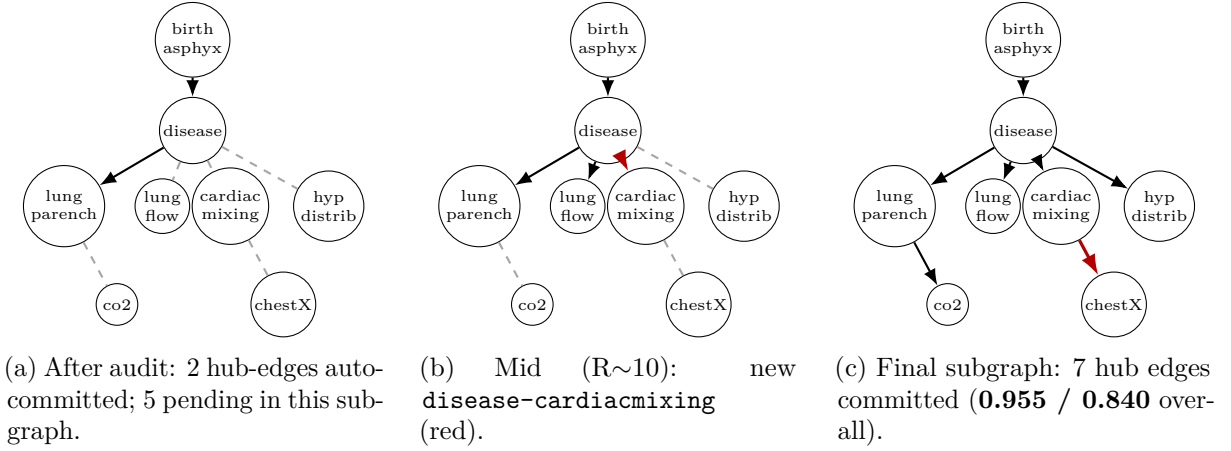

\centering
\begin{subfigure}[t]{0.31\linewidth}
  \centering
  \childdag{
    \draw[committed] (bir) -- (dis);
    \draw[committed] (dis) -- (lp);
    \draw[pending]   (dis) -- (lf);
    \draw[pending]   (dis) -- (cm);
    \draw[pending]   (dis) -- (hd);
    \draw[pending]   (lp)  -- (co2);
    \draw[pending]   (cm)  -- (cxr);
  }
  \caption{After audit: 2 hub-edges auto-committed; 5 pending in this subgraph.}
\end{subfigure}\hfill
\begin{subfigure}[t]{0.31\linewidth}
  \centering
  \childdag{
    \draw[committed] (bir) -- (dis);
    \draw[committed] (dis) -- (lp);
    \draw[committed] (dis) -- (lf);
    \draw[newedge]   (dis) -- (cm);
    \draw[pending]   (dis) -- (hd);
    \draw[pending]   (lp)  -- (co2);
    \draw[pending]   (cm)  -- (cxr);
  }
  \caption{Mid (R$\sim$10): new \texttt{disease-cardiacmixing} (red).}
\end{subfigure}\hfill
\begin{subfigure}[t]{0.31\linewidth}
  \centering
  \childdag{
    \draw[committed] (bir) -- (dis);
    \draw[committed] (dis) -- (lp);
    \draw[committed] (dis) -- (lf);
    \draw[committed] (dis) -- (cm);
    \draw[committed] (dis) -- (hd);
    \draw[committed] (lp)  -- (co2);
    \draw[newedge]   (cm)  -- (cxr);
  }
  \caption{Final subgraph: 7 hub edges committed (\textbf{0.955 / 0.840} overall).}
\end{subfigure}
\caption{child walk-through, hub-node subgraph view ($8$ of $20$
  nodes shown). The hub is \texttt{disease}, with all major clinical
  symptoms downstream. Edges outside the hub-subgraph evolve in
  parallel; the full $V=20$ DAG is too dense for a static figure.}
\label{fig:child-walk}
\end{figure*}

\FloatBarrier

\label{sec:appendix:alarm-walk}%
The final case study is alarm, the largest of the four benchmarks
at $V=37$. The full alarm DAG has $V=37$ nodes; we show the
cardiovascular hub subgraph
(HISTORY~$\to$~LVFAILURE~$\to$~LVEDVOLUME~$\to$~STROKEVOLUME~$\to$~CO~$\to$~BP)
plus the heart-rate branch (HR~$\to$~CO). The cascade audit + Meek
propagation auto-commit $2$ edges, and $31$ oracle queries
complete the full DAG at $0.810$ precision and $0.739$ recall.

\newcommand{\alarmdag}[1]{%
\begin{tikzpicture}[
  every node/.style={align=center, font=\tiny},
  v/.style={draw, circle, minimum size=0.55cm, inner sep=1pt, font=\tiny},
  committed/.style={-{Latex[length=2.2mm,width=1.7mm]}, thick, black},
  newedge/.style={-{Latex[length=2.8mm,width=2.0mm]}, very thick, red!70!black},
  pending/.style={dashed, gray!70, thick}
]
  \node[v] (his)  at (0.0, 1.8) {history};
  \node[v] (lvf)  at (1.4, 1.8) {lv\\failure};
  \node[v] (lvd)  at (2.8, 1.8) {lved\\volume};
  \node[v] (sv)   at (2.8, 0.6) {stroke\\volume};
  \node[v] (co)   at (1.4, 0.6) {co};
  \node[v] (bp)   at (0.0, 0.6) {bp};
  \node[v] (hr)   at (0.0, -0.6) {hr};
  \node[v] (hrbp) at (1.4, -0.6) {hrbp};
  #1
\end{tikzpicture}}

\begin{figure*}[!htb]
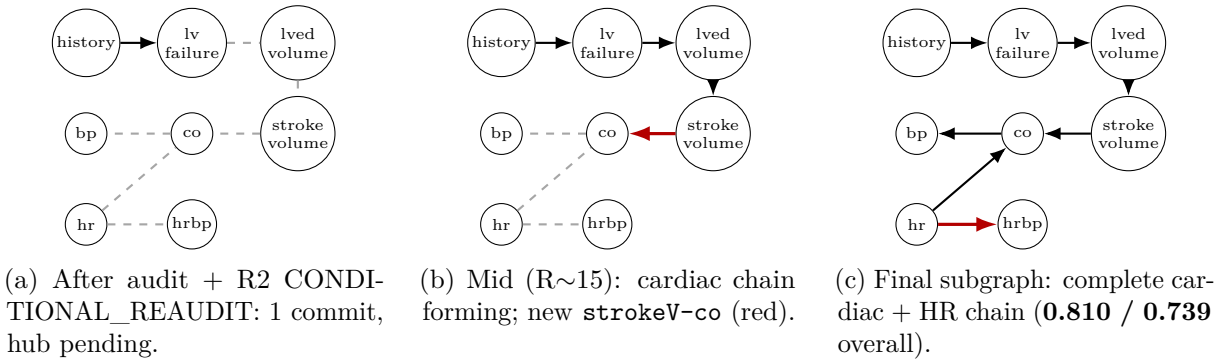

\centering
\begin{subfigure}[t]{0.31\linewidth}
  \centering
  \alarmdag{
    \draw[committed] (his) -- (lvf);
    \draw[pending]   (lvf) -- (lvd);
    \draw[pending]   (lvd) -- (sv);
    \draw[pending]   (sv)  -- (co);
    \draw[pending]   (co)  -- (bp);
    \draw[pending]   (hr)  -- (co);
    \draw[pending]   (hr)  -- (hrbp);
  }
  \caption{After audit + R2 CONDITIONAL\_REAUDIT: 1 commit, hub pending.}
\end{subfigure}\hfill
\begin{subfigure}[t]{0.31\linewidth}
  \centering
  \alarmdag{
    \draw[committed] (his) -- (lvf);
    \draw[committed] (lvf) -- (lvd);
    \draw[committed] (lvd) -- (sv);
    \draw[newedge]   (sv)  -- (co);
    \draw[pending]   (co)  -- (bp);
    \draw[pending]   (hr)  -- (co);
    \draw[pending]   (hr)  -- (hrbp);
  }
  \caption{Mid (R$\sim$15): cardiac chain forming; new \texttt{strokeV-co} (red).}
\end{subfigure}\hfill
\begin{subfigure}[t]{0.31\linewidth}
  \centering
  \alarmdag{
    \draw[committed] (his) -- (lvf);
    \draw[committed] (lvf) -- (lvd);
    \draw[committed] (lvd) -- (sv);
    \draw[committed] (sv)  -- (co);
    \draw[committed] (co)  -- (bp);
    \draw[committed] (hr)  -- (co);
    \draw[newedge]   (hr)  -- (hrbp);
  }
  \caption{Final subgraph: complete cardiac + HR chain (\textbf{0.810 / 0.739} overall).}
\end{subfigure}
\caption{alarm walk-through, cardiovascular-hub subgraph view ($8$
  of $37$ nodes shown). The hub is the cardiac output chain;
  side branches (e.g., respiratory) evolve in parallel. The full
  $V=37$ DAG is too dense to render statically; the trace CSV
  records the complete $31$-round event sequence.}
\label{fig:alarm-walk}
\end{figure*}

\FloatBarrier

\end{document}